\documentclass[A4paper, 11 pt, conference]{ieeeconf}  
\IEEEoverridecommandlockouts       
\usepackage{geometry}
\usepackage{epsfig} % for postscript graphics files
\usepackage{times} % assumes new font selection scheme installed
\usepackage{amsmath} % assumes amsmath package installed
\usepackage{amssymb}  % assumes amsmath package installed
\usepackage{algorithm} %format of the algorithm
\usepackage{algorithmic} %format of the algorithm
\usepackage{multirow} %multirow for format of table

\usepackage{graphicx}
\usepackage[subfigure]{graphfig} 
\usepackage{stfloats}
\usepackage{subeqnarray}
\usepackage{cases}
\usepackage{comment}

\begin{document}

% Change to your title
\title{\LARGE \bf
Prioritized Hierarchical Compliance Control for \\ Dual-Arm Robot Stable Clamping}

\author{Xiaoyu Ren$^{1}$, Liqun Huang$^{1}$ and Mingguo Zhao$^{2}$% <-this % stops a space
	%\thanks{*This work was supported by Ubtech Robotics Corp.}% <-this % stops a space
	\thanks{$^{1}$Xiaoyu Ren and Liqun Huang are with Ubtech Robotics Corp., Beijing, China. Email: xiaoyu007.ren@ubtrobot.com, litchi.huang@ubtrobot.com.}%
	\thanks{$^{2}$Mingguo Zhao is the corresponding author and with Department of Automation, Tsinghua University and Beijing Innovation Center for Future Chips, Tsinghua University, Beijing, China. Email: mgzhao@mail.tsinghua.edu.cn.}}

\maketitle 
\thispagestyle{empty}

\begin{abstract}
When a dual-arm robot clamps a rigid object in an environment for human beings, the environment or the collaborating human will impose incidental disturbance on the operated object or the robot arm, leading to clamping failure, damaging the robot even hurting the human. This research proposes a prioritized hierarchical compliance control to simultaneously deal with the two types of disturbances in the dual-arm robot clamping. First, we use hierarchical quadratic programming (HQP) to solve the robot inverse kinematics under the joint constraints and prioritize the compliance for the disturbance on the object over that on the robot arm. Second, we estimate the disturbance forces throughout the momentum observer with the F/T sensors and adopt admittance control to realize the compliances. Finally, we perform the verify experiments on a 14-DOF position-controlled dual-arm robot WalkerX, clamping a rigid object stably while realizing the compliance against the disturbances.
\end{abstract}

\section{Introduction}
The dual-arm robot has more significant advantages in dexterity, load capacity, and adaptability \cite{2018Dual}\cite{2017Indirect}. They are often more suitable for complex operation tasks in application scenarios such as restaurants, shopping malls, and airports. Simultaneously, completing these complex operation tasks by dual-arm robots in an environment suitable for human beings will be subject to unpredictable interference from the surrounding objects and human beings.  Hence, we concentrate on the control of dual-arm clamping under unpredictable external disturbances in this paper. We divide the external disturbances into two types according to the location where it is applied, to the operated object, and the robot arm.

The disturbance on the object may cause excessive interaction force and break the contact and friction constraints, leading to the clamping failure. The admittance/impedance control on the operated object was commonly used to comply with the disturbance \cite{bjerkeng2014fast}\cite{Ren2016Biomimetic}.
The internal forces that do not contribute to the object motion were usefully exploited to maintain a firm clamping against the disturbance \cite{murray1994mathematical}.
A common strategy was taken to regulate the internal impedance between the end-effectors and the object \cite{caccavale2008six}\cite{heck2013internal}\cite{2016Coordinated}\cite{han2019collision} or the impedance of the relative task \cite{tarbouriech2019admittance} when using the cooperative task-space representation \cite{chiacchio1996direct}. 
Recent works \cite{lin2018projected}\cite{gao2018projected} utilized the projected inverse dynamics to decompose the contact force and realize the explicit force control in the constrained subspace without generating extra object motion.  
The target contact forces can be obtained by setting a safety threshold of friction \cite{2014Dual} or building an optimization problem to ensure the friction constraints \cite{lin2018projected}\cite{2007Fast}. 
Considering the extreme cases that all tasks could not be executed wholly,  in \cite{tarbouriech2019admittance}, priority is given  to the control of internal forces to ensure the firm contact first.  

These works rarely considered the disturbance on the robot arm, such as a sudden collision, which can easily damage the robot and even injure humans. Thus the activated impedance behavior was usually imposed in the joint space to ensure the safe interaction \cite{capurso2017sensorless}. 
However, it would affect the correct execution of the main tasks, such as clamping an object.
For redundant robots, the null-space projection has been widely adopted to ensure the joint space compliance behavior did not interfere with the main tasks \cite{2012Null}\cite{2014Task}\cite{2018Control}. However, the null-space method was difficult to satisfy the joint constraints, and the processing of the inequality tasks was very complex and inefficient \cite{flacco2012prioritized}\cite{flacco2012motion}.

The above solutions could handle the two types of external disturbances separately. However, when handling disturbances simultaneously, they will affect each other, destroy clamping stability or easily exceed joint constraints. 
In this paper, we propose a prioritized hierarchical compliance control approach for dual-arm robot clamping.    
Our approach is as follows:
the momentum observer \cite{4058607}\cite{2018A} and the F/T sensors data are exploited to decouple two-type disturbances in the joint space. Then they are fed into admittance controllers to realize compliance subject to the disturbances. We construct an optimization problem with friction cone constraints to keep the clamping stable even when suffering from external disturbances. The HQP is used to ensure that the joint space compliance behavior does not disturb the clamping tasks. The contributions of this paper are proposing a prioritized hierarchical compliance control approach to realize stable clamping while resisting disturbances on the operated object and the robot arms simultaneously, and deploying it in a position-controlled dual-arm robot with a real-time cycle of 1ms. 

\section{Method}
\begin{figure*}[btp]
	\centering
	\includegraphics[width=17cm]{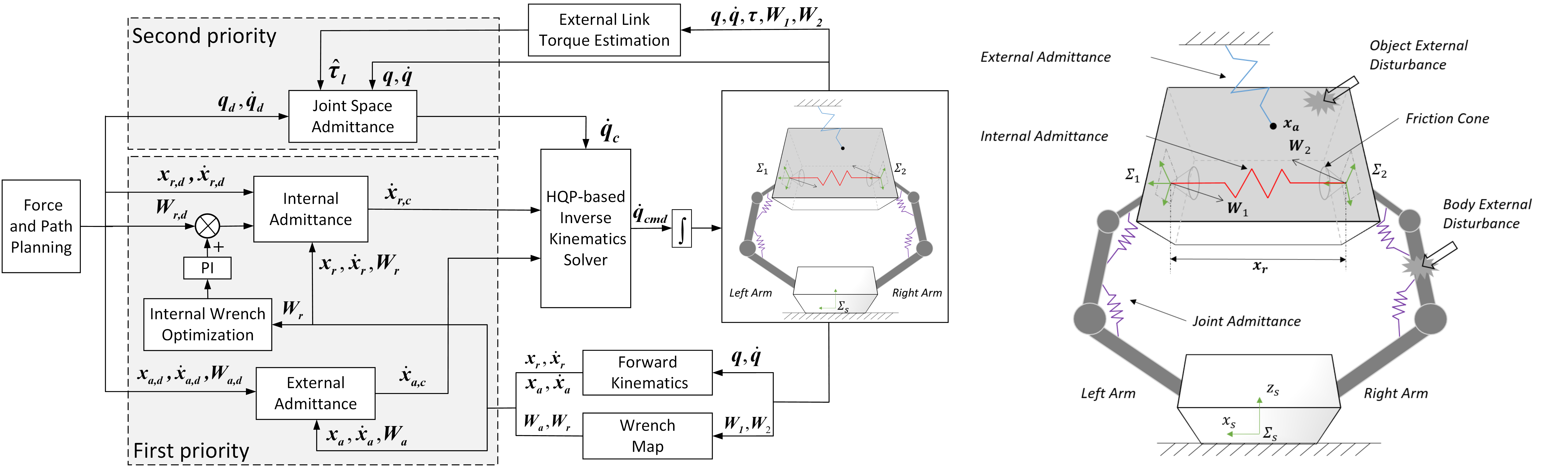}
	\caption{Control framework and clamping model of the dual-arm robot.}
	\label{dual arm clamp framework}
\end{figure*}

The proposed approach is based on a two-level HQP, together with admittance controllers for the body and object disturbances, and the optimization and force feedback control for the internal object wrench. An overview of the control framework is illustrated in Fig.\ref{dual arm clamp framework}.
\subsection{System Modeling}\label{object control}
%Kinematics and Dynamics of dual-arm robot 
As shown in Fig.\ref{dual arm clamp framework} (b), the following coordinate frames are used to describe the dual-arm robot system with two $n$-DOF arms: $\varSigma_{s}$ is the base coordinate frame , $\varSigma_{1}$ and $\varSigma_{2}$ are the left and right coordinate frames attached to each contact point of end-effectors, and $\varSigma_{a}$ is a frame fixed on the object. The equations of motion in the joint space can be given by  
\begin{equation}
	\boldsymbol{M}(\boldsymbol{q})\ddot{\boldsymbol{q}} + \boldsymbol{C}(\boldsymbol{q},\dot{\boldsymbol{q}}) \dot{\boldsymbol{q}} + \boldsymbol{G}(\boldsymbol{q}) +  \boldsymbol{\tau}_{f}(\boldsymbol{q},\dot{\boldsymbol{q}}) = \boldsymbol{\tau} + \boldsymbol{\tau}_{ext}
	\label{dynamics}
\end{equation}	
where $\boldsymbol{q} = \left[\boldsymbol{q}_{1}^{\mathit{T}}, \boldsymbol{q}_{2}^{\mathit{T}}\right]^{\mathit{T}} \in \mathbb{R}^{2n}$ is the generalized coordinates. $\boldsymbol{\tau}, {\boldsymbol{\tau}}_{f}, \boldsymbol{\tau}_{ext} \in \mathbb{R}^{2n}$ are the actuation torques, friction torques,  the external joint torques resulting from the possible external wrenches. $\boldsymbol{M}, \boldsymbol{C}, \boldsymbol{G}$ are the inertia matrix, Coriolis and centrifugal matrix, the gravity vector. 

When the robot clamps an object rigidly without slipping, the external joint torques can be calculated from
\begin{equation}
	\boldsymbol{\tau}_{ext} = -\boldsymbol{J}_{c}^{\mathit{T}} \boldsymbol{W}_{c}
	\label{contact wrench}
\end{equation}
where $\boldsymbol{W}_{c} = \left[\boldsymbol{W}_{1}^{\mathit{T}}, \boldsymbol{W}_{2}^{\mathit{T}}\right]^{\mathit{T}} \in \mathbb{R}^{12} $ and $\boldsymbol{W}_{i} = \left[\boldsymbol{m}_{i}^{\mathit{T}}, \boldsymbol{f}_{i}^{\mathit{T}} \right]^{\mathit{T}} \in \mathbb{R}^{6}$ is the $i^{th}$ contact wrench applied to the object expressed in base frame, $\boldsymbol{m}_{i}, \boldsymbol{f}_{i}$ denote the contact moment and force. $ \boldsymbol{J}_{c} = \left[\boldsymbol{J}_{1}^{\mathit{T}}, \boldsymbol{J}_{2}^{\mathit{T}}\right]^{\mathit{T}}  \in \mathbb{R}^{12 \times 2n}$ and $\boldsymbol{J}_{i} \in \mathbb{R}^{6 \times 2n}$ is the $i^{th}$ contact Jacobian matrix. Note that, if not specfied, all variables in the rest of this paper are expressed in the base frame and for simplicity we eliminate the superscript. 

%Grasp map
For the convenience of understanding the effect of the contact wrenches on the object, the total object wrench and the internal object wrench are considered. Referring to the definition in \cite{murray1994mathematical}, the total object wrench due to the dual-arm clamping, which has an influence on the object motion, can be constructed as
\begin{equation}
	\boldsymbol{W}_{a} = \begin{bmatrix}
		{}^a\!\boldsymbol{X}_{1}^{*} & {}^a\!\boldsymbol{X}_{2}^{*}
	\end{bmatrix}\begin{bmatrix}
		\boldsymbol{W}_{1} \\
		\boldsymbol{W}_{2}
	\end{bmatrix}=\boldsymbol{G} \boldsymbol{W}_{c}
	\label{total object wrench}
\end{equation}
where ${}^a\!\boldsymbol{X}_{i}^{*} \in \mathbb{R}^{6 \times 6}$ is the wrench transformation matrix \cite{2008Rigid} which transforms contact wrenches to the object frame, \(\boldsymbol{G}\) is the grasp map which defines the map between the contact wrenches and the total object wrench.

The internal object wrench, which causes no net object wrench, can ensure that the contact wrenches satisfy friction constraints. The internal wrench can be constructed in more than one way. For example, when constructed as \cite{uchiyama1988symmetric}, the decoupled internal object wrench with two contact points can be described as 
\begin{equation}
	\boldsymbol{W}_{r} = \begin{bmatrix}
		-\frac{1}{2}\boldsymbol{I} & \frac{1}{2}{}^1\!\boldsymbol{X}_{2}^{*}
	\end{bmatrix} \begin{bmatrix}
		\boldsymbol{W}_{1} \\ \boldsymbol{W}_{2}
	\end{bmatrix} = \bar{\boldsymbol{E}} \boldsymbol{W}_{c}
	\label{internal wrench}
\end{equation}
where $\boldsymbol{I}$ denotes a $6 \times 6$ identity matrix, ${}^1\!\boldsymbol{X}_{2}^{*} \in \mathbb{R}^{6 \times 6}$ transforms the wrench to the frame $\varSigma_{1}$, $\bar{\boldsymbol{E}} \in \mathbb{R}^{6 \times 12}$ is the mapping matrix between the contact wrenches and the internal object wrench. Combined (\ref{total object wrench}) with (\ref{internal wrench}), we can get 
\begin{equation}
	\begin{aligned}
		\begin{bmatrix}
			\boldsymbol{W}_a\\\boldsymbol{W}_r
		\end{bmatrix} &=\begin{bmatrix}
			\boldsymbol{G}\\ \bar{\boldsymbol{E}}
		\end{bmatrix}\begin{bmatrix}
			\boldsymbol{W}_{1} \\\boldsymbol{W}_{2}
		\end{bmatrix} 
		&=\boldsymbol{Q}\begin{bmatrix}
			\boldsymbol{W}_{1} \\\boldsymbol{W}_{2}
		\end{bmatrix}
	\end{aligned}\label{g_5}
\end{equation}
where \(\boldsymbol{Q}\) maps contact wrenches to the total object wrench and internal object wrench. Due to the virtual work principle, the contact kinematics has the form as (\ref{g_8}),
\begin{equation}
	\begin{aligned}
		\begin{bmatrix}
			\boldsymbol{\dot{x}}_a\\
			\boldsymbol{\dot{x}}_r
		\end{bmatrix} =  \boldsymbol{Q}^{-T} \begin{bmatrix}
			\boldsymbol{\dot{x}}_1\\
			\boldsymbol{\dot{x}}_2
		\end{bmatrix}
	\end{aligned}
	\label{g_8}
\end{equation}
where \(\boldsymbol{\dot{x}}_a, \boldsymbol{\dot{x}}_r \in \mathbb{R}^6\) denote the absolute velocity of the object and the relative velocity of the two contact points, $\boldsymbol{ \dot{x}}_i = \boldsymbol{J}_{i} \boldsymbol{\dot{q}}_i \in \mathbb{R}^6$  is the velocity of each contact point. 

\subsection{Controller}
1) \textit{Admittance control for object wrench}

When the object clamped by the dual-arm robot suffers from the external disturbance on it, two independent controllers are designed to resist the disturbance exerted on the object and regulate the contact wrenches to avoid slipping, respectively. The dynamics equation of the object can be described as
\begin{equation}
	\boldsymbol{W}_{ext} + \boldsymbol{W}_{a} + \boldsymbol{G}_{a} = \boldsymbol{M}_{a}  \boldsymbol{\ddot{x}}_{a} + \begin{bmatrix}
	    \boldsymbol{\omega}_{a} \times (\boldsymbol{I}_a \boldsymbol{\omega}_{a}) \\
	    \boldsymbol{0}
	\end{bmatrix}
	\label{gg_1}
\end{equation}
where $\boldsymbol{W}_{ext} \in \mathbb{R}^6$ is the disturbance wrench on the object, $\boldsymbol{G}_{a} \in \mathbb{R}^6$ is gravity vector of the object, $\boldsymbol{M}_{a} \in \mathbb{R}^{6 \times 6}$ is the inertial matrix of the object, $\boldsymbol{I}_a \in \mathbb{R}^{3 \times 3}$ is the inertia tensor, $\boldsymbol{\omega}_{a} \in \mathbb{R}^{3}$ is the angular velocity of the object. Then the Cartesian admittance control is used to describe the disturbance response of the object. Right side of (\ref{gg_1}) is approximate to $0$ due to the slow movement and the constant orientation of the object. Gravity is regarded as a disturbance because of the unknown mass of the clamped object. (\ref{g_17}) gives the control law, where $\tilde{\boldsymbol{x}}_{a} = \boldsymbol{x}_{a} - \boldsymbol{x}_{a,d}$, $\boldsymbol{x}_{a,d} \in SE(3)$ is the desired equilibrium point, $\boldsymbol{\dot{x}}_{a,c} \in \mathbb{R}^6$ is the command object velocity,   \(\boldsymbol{K}_{a},\boldsymbol{B}_{a} \in \mathbb{R}^{6 \times 6}\) are positive stiffness and damping matrices, respectively.
\begin{equation}
	\begin{aligned}
		-\boldsymbol{W}_{a} &= \boldsymbol{K}_{a} \tilde{\boldsymbol{x}}_{a} + \boldsymbol{B}_{a} \dot{\tilde{ \boldsymbol{x}}}_{a} \\
		\boldsymbol{\dot{x}}_{a,c} &= \dot{\boldsymbol{x}}_{a,d} + \boldsymbol{B}_{a}^{-1}(-\boldsymbol{W}_{a} - \boldsymbol{K}_{a} \tilde{\boldsymbol{x}}_{a})
	\end{aligned}
	\label{g_17}
\end{equation}

2) \textit{Admittance control for optimal internal object wrench }

When there are arbitrary disturbances exerted on the object or the end-effectors, the contact wrenches may be out of friction cone, leading to the rolling and slippage on the surface of the object. As stated above, a sufficient enough internal object wrench can be chosen to regulate the contact wrenches to remain in the friction cone without affecting the total object wrench. As shown in Fig.\ref{dual arm clamp framework} (b), the two end-effectors with rectangle blocks have face contacts with the object and the $x$-axis is normal to the contact plane. So the optimization problem with the range constraints has the form as \eqref{optf}. (\ref{optf a}) and (\ref{optf b}) are the constraints of friction cones. (\ref{optf c}) and (\ref{optf d}) are to make sure the center of pressure to be in the contact patch. Note that in practice we approximate the circle friction cone by a octagon cone.  
\begin{subequations}
	\begin{align}
		\boldsymbol{W}_{r,dx}^{*} =& \underset{\boldsymbol{W}_{r,x}} {\operatorname{argmin}} \quad \lVert \boldsymbol{W}_{r,x} \rVert_{2} \\
		s.t. \qquad & \boldsymbol{W}_{r} = \bar{\boldsymbol{E}}\begin{bmatrix}
			\boldsymbol{W}_{1} \\ \boldsymbol{W}_{2}
		\end{bmatrix} \\
		& \boldsymbol{W}_{i,x} \leq 0, \quad i = \{1,2\} \\
		& \sqrt{\boldsymbol{W}_{i,y}^{2} + \boldsymbol{W}_{i,z}^{2}} \leq \mu_{i} | \boldsymbol{W}_{i,x}| \label{optf a} \\
		& | \boldsymbol{W}_{i,mx} | \leq \lambda_{i} |\boldsymbol{W}_{i,x} | \label{optf b} \\
		& | \boldsymbol{W}_{i,my} | \leq \frac{1}{2} \delta_{i,z} |\boldsymbol{W}_{i,x} | \label{optf c} \\
		& | \boldsymbol{W}_{i,mz} | \leq \frac{1}{2} \delta_{i,y} |\boldsymbol{W}_{i,x}| \label{optf d}
	\end{align}
	\label{optf} 
\end{subequations}
where the subscript of $\{x,y,z, mx, my, mz\}$ denote the forces and moments along each axis, the subscript of $\{d\}$ denotes the desire one. \(\mu_{i}, \lambda_{i}\) are the tangential and torsional friction coefficients. \(\delta_{i,y}, \delta_{i,z}\) are the side lengths of two end-effectors along $y$-axis and $z$-axis.  

To realize a good tracking effect on the optimal internal object wrench $\boldsymbol{W}_{r,dx}^{*}$, we create a simple PID-controller as (\ref{pid}) to get the resultant $\boldsymbol{W}^{*}_{r,d}$ for the internal admittance control. Here the velocity acts as feedback instead of the noisy differential wrench for robustness.
\begin{equation}
	\boldsymbol{W}^{*}_{r,d} = \boldsymbol{W}_{r,d} +  \boldsymbol{S}_x \left(k_p \tilde{\boldsymbol{f}}_x + k_i \int \tilde{\boldsymbol{f}}_x dt - k_d \dot{\boldsymbol{x}}_{r,x} \right)
	\label{pid}
\end{equation}
where $\boldsymbol{W}_{r,d} \in \mathbb{R}^{6 \times 1}$ is the desired internal object wrench, $\tilde{\boldsymbol{f}}_x = \boldsymbol{W}_{r,dx}^{*} - \boldsymbol{W}_{r,x}$ is the wrench error, \(\boldsymbol{S}_x=[0, 0, 0, 1, 0, 0]^{\mathit{T}}\) is a selection vector of $x$-axis normal to the contact plane, and $k_p,k_i,k_d$ are the PID feedback gains. Therefore the control law for internal admittance control is given as  
\begin{equation}
	\begin{aligned}
		\tilde{\boldsymbol{W}}_{r}
		&=\boldsymbol{K}_{r} \tilde{\boldsymbol{x}}_{r} + 
		\boldsymbol{B}_{r} \dot{\tilde{\boldsymbol{x}}}_{r} \\
		\boldsymbol{\dot{x}}_{r,c} &= \dot{\boldsymbol{x}}_{r,d} + \boldsymbol{B}_{r}^{-1}(\tilde{\boldsymbol{W}}_{r} - \boldsymbol{K}_{r} \tilde{\boldsymbol{x}}_{r})
	\end{aligned}
	\label{g_18}
\end{equation}
where $\tilde{\boldsymbol{x}}_{r} = \boldsymbol{x}_{r} - \boldsymbol{x}_{r,d}$, $\boldsymbol{x}_{r,d} \in SE(3)$ is the desired equilibrium point, $\boldsymbol{\dot{x}}_{r,c} \in \mathbb{R}^{6}$ is the command relative velocity, $\tilde{\boldsymbol{W}}_{r} = \boldsymbol{W}_{r} - \boldsymbol{W}^{*}_{r,d}$ is the wrench error,  \(\boldsymbol{K}_{r},\boldsymbol{B}_{r} \in \mathbb{R}^{6 \times 6} \) are the positive stiffness and the damping matrices.  

3) \textit{Admittance control for disturbances on robot arm}

During task execution, the robot may suffer from the sudden disturbances on both robot arms in addition to the disturbance on the object. At this moment, all the disturbances contribute to the external joint torques and we need to rewrite (\ref{contact wrench}) as 
\begin{equation}
	\boldsymbol{\tau}_{ext} = -\boldsymbol{J}_{c}^{\mathit{T}} \boldsymbol{W}_{c} + \boldsymbol{\tau}_{l}
\end{equation}
where $\boldsymbol{\tau}_{l} = \sum^{m}_{k = 1} \boldsymbol{J}_{l,k}^\mathit{T} \boldsymbol{W}_{l,k} \in \mathbb{R}^{2n}$ is the total external joint torques caused by disturbances on the robot arms, $\boldsymbol{J}_{l,k} \in \mathbb{R}^{6 \times 2n}, \boldsymbol{W}_{l,k} \in \mathbb{R}^{6}$ are the $k^{th}$ contact Jacobian matrix and the contact wrench. 

Because the number, location and amplitude of contacts on the arms are random and usually unknown without additional perception, it is a better choice to comply with disturbances in the joint space rather than in the Cartesian space. So we construct a generalized momentum observer to estimate the total external joint torques $\boldsymbol{\tau}_{ext}$. The skew symmetric matrix $\dot{\boldsymbol{M}}-2\boldsymbol{C}$ is utilized to eliminate the joint accelerations. The contact wrenches at the end-effectors can be easily obtained by the 6-axis Force/Torque sensors, thus we can obtain the estimated $\boldsymbol{\tau}_{l}$ for joint admittance control as 
\begin{equation}
	\begin{aligned}
		\hat{\boldsymbol{\tau}}_{ext} &= 
		\boldsymbol{K} (\boldsymbol{M} \boldsymbol{\dot{q}} - \int( \boldsymbol{\tau} + \boldsymbol{C}^T \dot{\boldsymbol{q}}
		- \boldsymbol{G} - {\boldsymbol{\tau}}_{f} + \hat{\boldsymbol{\tau}}_{ext}) dt ) \\
		\hat{\boldsymbol{\tau}}_{l} &= \hat{\boldsymbol{\tau}}_{ext} + \boldsymbol{J}_{c}^{\mathit{T}} \boldsymbol{W}_{c} 
	\end{aligned}	
	\label{2.10}
\end{equation}
where $\boldsymbol{K} \in \mathbb{R}^{2n \times 2n} $ is the diagonal and positive gain matrix. The estimated $\hat{\boldsymbol{\tau}}_{ext}$ and $\hat{\boldsymbol{\tau}}_{l}$ will gradually converge to the actual ones. Finally the admittance control law for $\hat{\boldsymbol{\tau}}_{l}$ is given as
\begin{equation}
	\begin{aligned}
		\hat{\boldsymbol{\tau}}_{l} &= \boldsymbol{\Lambda}_{j} \ddot{ \tilde{ \boldsymbol {q}}} +\boldsymbol{B}_{j} \dot{ \tilde{ \boldsymbol{q}}} + \boldsymbol{K}_{j} \tilde{\boldsymbol{q}} \\
		\dot{\boldsymbol{q}}_{c} &= \dot{\boldsymbol{q}}_{d} +  \int \left(\ddot{\boldsymbol{q}}_{d}+\boldsymbol{\Lambda}_{j}^{-1}(\hat{\boldsymbol{\tau}}_{l} - \boldsymbol{B}_{j} \dot{ \tilde{ \boldsymbol{q}}} - \boldsymbol{K}_{j} \tilde{\boldsymbol{q}}) \right)dt
	\end{aligned}
	\label{JAD}
\end{equation}
where $\tilde{\boldsymbol{q}} = \boldsymbol{q} - \boldsymbol{q}_d$, $\boldsymbol{q}_d \in \mathbb{R}^{2n}$ is the desired equilibrium point, $\dot{\boldsymbol{q}}_{c} \in \mathbb{R}^{2n}$ is the command velocity of joint space, $\boldsymbol{\Lambda}_{j}, \boldsymbol{B}_{j}$, $\boldsymbol{K}_{j} \in \mathbb{R}^{2n \times 2n}$ are the positive inertia, damping and stiffness matrices, respectively.

4) \textit{Prioritized hierarchical control frame}

In the real world, the dual-arm robot is usually asked to execute multiple tasks simultaneously rather than a single task at a time. Moreover, it is important to maintain the joint constraints, such as the limits of position, velocity and torque. For the robot clamping an object, the tasks are considered: clamping the object stably, resisting the disturbances on the object and resisting the disturbances on the robot arms. Without violating the joint constraints, the first two tasks should be more important than the joint space compliance when tasks are conflict with each other and can not be executed completely. So the task priority order for the dual arm clamping control is 

joint limits \(\succ\) stable clamping = disturbances on the object \(\succ\) disturbances on the robot arms. 

To solve the hierarchical optimal control with inequality constraints, the hierarchical quadratic programming (HQP) is used to solve the inverse kinematics and find the optimal command velocity. The first optimal problem is to realize the admittance control for the object wrench and the optimal internal wrench as far as possibly.
\begin{equation}
	\begin{aligned}
		\dot{\boldsymbol{q}}_{1st}^{*} &=\underset{\dot{q}}{\arg \min } \left\| \boldsymbol{J}_{c} \dot{\boldsymbol{q}} - \dot{\boldsymbol{x}}_{cmd} \right\|_{2} + \rho \|\dot{\boldsymbol{q}}\|_{2} \\
		\text { s.t. } &\dot{\boldsymbol{x}}_{cmd} = \boldsymbol{Q}^\mathit{T} \begin{bmatrix}
			\dot{\boldsymbol{x}}_{a,c} \\
			\dot{\boldsymbol{x}}_{r,c}
		\end{bmatrix} \\
		&\boldsymbol{b}_{m} \leq \dot{\boldsymbol{q}} \leq \boldsymbol{b}_{M} \\
		&\boldsymbol{b}_{m} = \operatorname{max}(\dot{\boldsymbol{q}}_{min}, \left(\boldsymbol{q}_{min} - \boldsymbol{q}\right)/\Delta t) \\
		&\boldsymbol{b}_{M} = \operatorname{min}(\dot{\boldsymbol{q}}_{max}, \left(\boldsymbol{q}_{max} - \boldsymbol{q}\right)/\Delta t)
	\end{aligned}
	\label{first QP}
\end{equation}
where $\rho$ is the damping term added to avoid the abnormal joint velocity in a singular configuration, $\boldsymbol{q}_{min}, \boldsymbol{q}_{max} \in \mathbb{R}^{2n}$ are the bounds of joint positions, $\dot{\boldsymbol{q}}_{min}, \dot{\boldsymbol{q}}_{max} \in \mathbb{R}^{2n}$ are the bounds of joint velocities, $\Delta t$ is the time interval. 

The second optimal problem is to realize the admittance control for external joint torques as far as possible without affecting the result of the first QP. So the optimal value \( \dot{\boldsymbol{q}}_{1st}^{*} \) is regarded as one of the constraints in the second QP. The final control velocity in the joint space, which can realize the three tasks hierarchically, can be computed as 
\begin{equation}
	\begin{aligned}
		\dot{\boldsymbol{q}}_{cmd} &=\underset{\dot{q}}{\arg \min }\left\|\dot{\boldsymbol{q}}-\dot{\boldsymbol{q}}_{c}\right\|_{2} \\
		\text {s.t. } &\boldsymbol{J}_{c} \dot{\boldsymbol{q}}=\boldsymbol{J}_{c} \dot{\boldsymbol{q}}_{1st}^{*} \\
		& \boldsymbol{b}_{m} \leq \dot{\boldsymbol{q}} \leq \boldsymbol{b}_{M}
	\end{aligned}
	\label{second qp}
\end{equation}

\section{Validation}
The proposed control approach is verified experimentally in a dual-arm robot with two 7-DOF robot arms. The robot is position-controlled, with position sensors in each joint and F/T sensors at the end effectors. The joint torques are obtained from the joint currents and the current-to-torque factors. The optimization problems (\ref{optf}) (\ref{first QP}) (\ref{second qp}) are solved using the open-source software package qpOASES \cite{qpOASES2017}. The controller runs at a control cycle of 1 ms on the Core i7-7600U CPU. The safe friction coefficients for the internal wrench optimization are preset to \(\mu=0.15\) and \(\lambda= 0.01\) respectively according to the actual ones \(\mu^*=0.2\) and \(\lambda^*= 0.0133\).
\begin{figure}[tbhp]
	\centering
	\subfigtopskip=2pt 
	\subfigbottomskip=2pt 
	\subfigcapskip=-2pt 
	\subfigure[]{
		\centering
		\includegraphics[width=8.3cm]{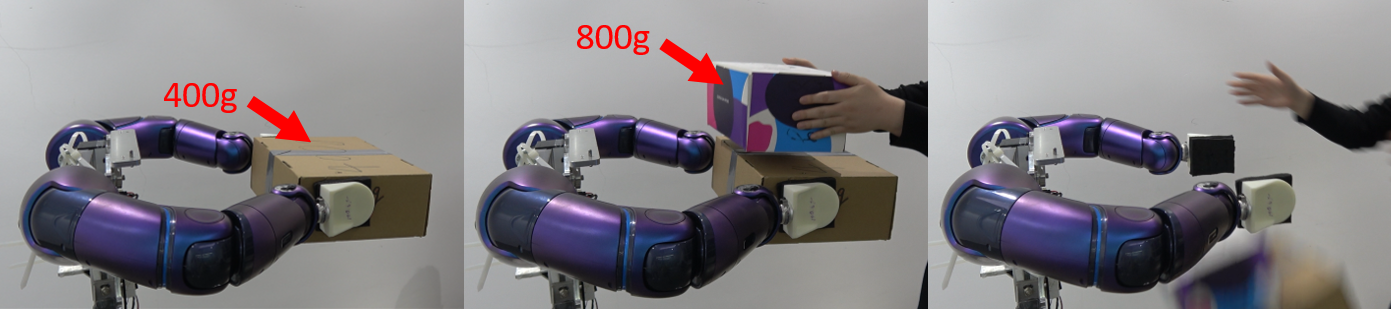} 
	}
	\quad
	\subfigure[]{
		\centering
		\includegraphics[width=8.3cm]{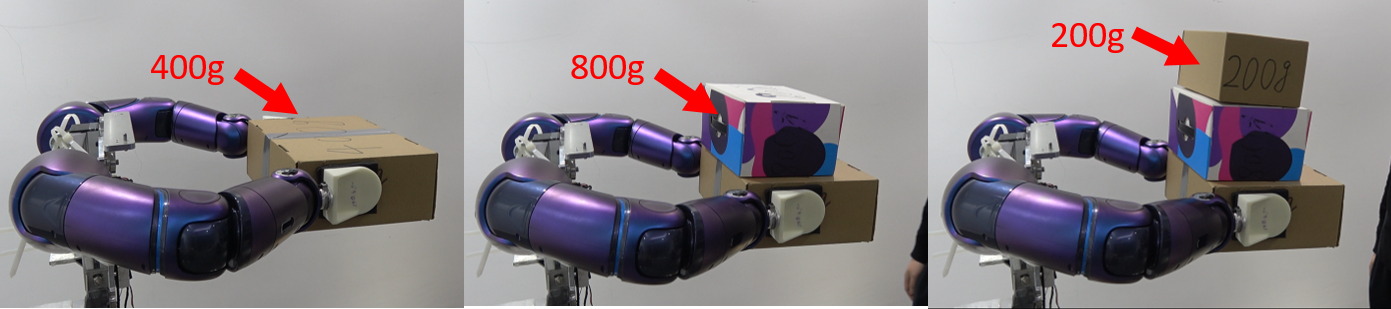}
	}
	\caption{Keep a constant internal wrench (a). Track an optimized internal wrench (b).}
	\label{exp1}
\end{figure}

\begin{figure}[tbhp]
	\centering
	\subfigtopskip=2pt 
	\subfigbottomskip=2pt 
	\subfigcapskip=-2pt 
	\subfigure[]{
		\centering
		\includegraphics[width=8.3cm]{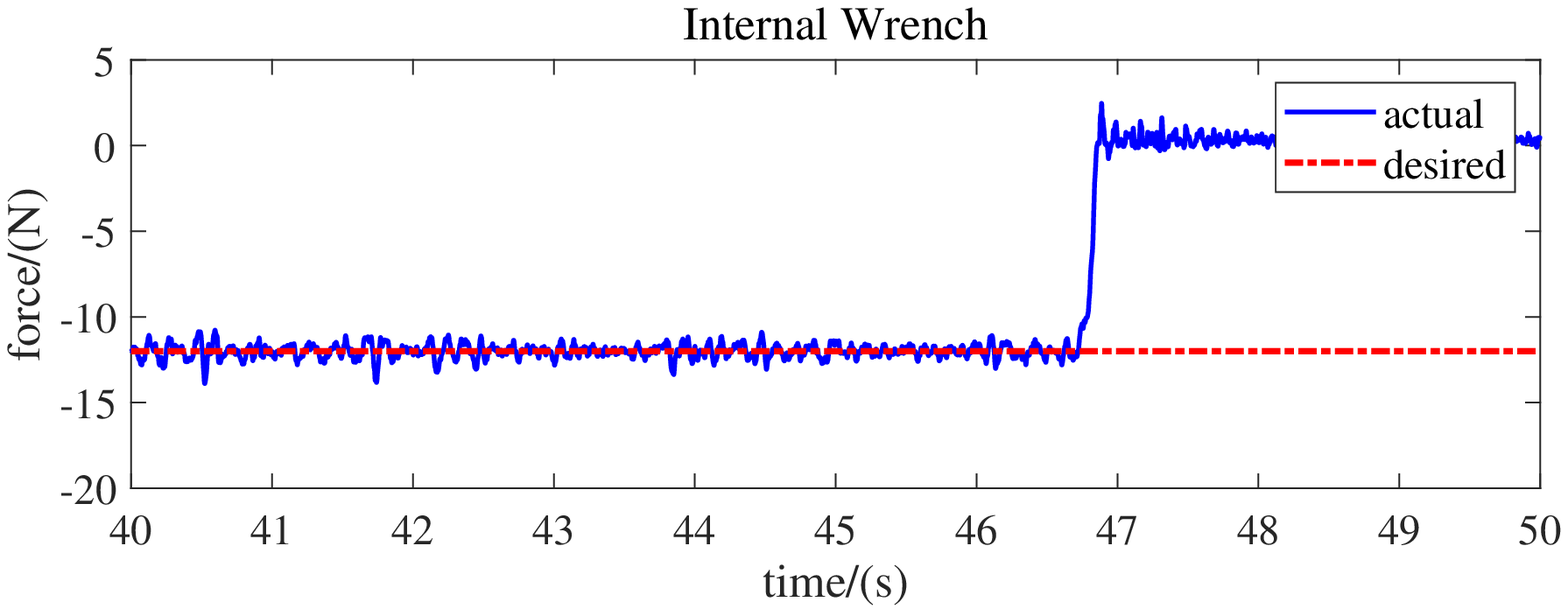}
	}
	
	\subfigure[]{
		\centering
		\includegraphics[width=8.3cm]{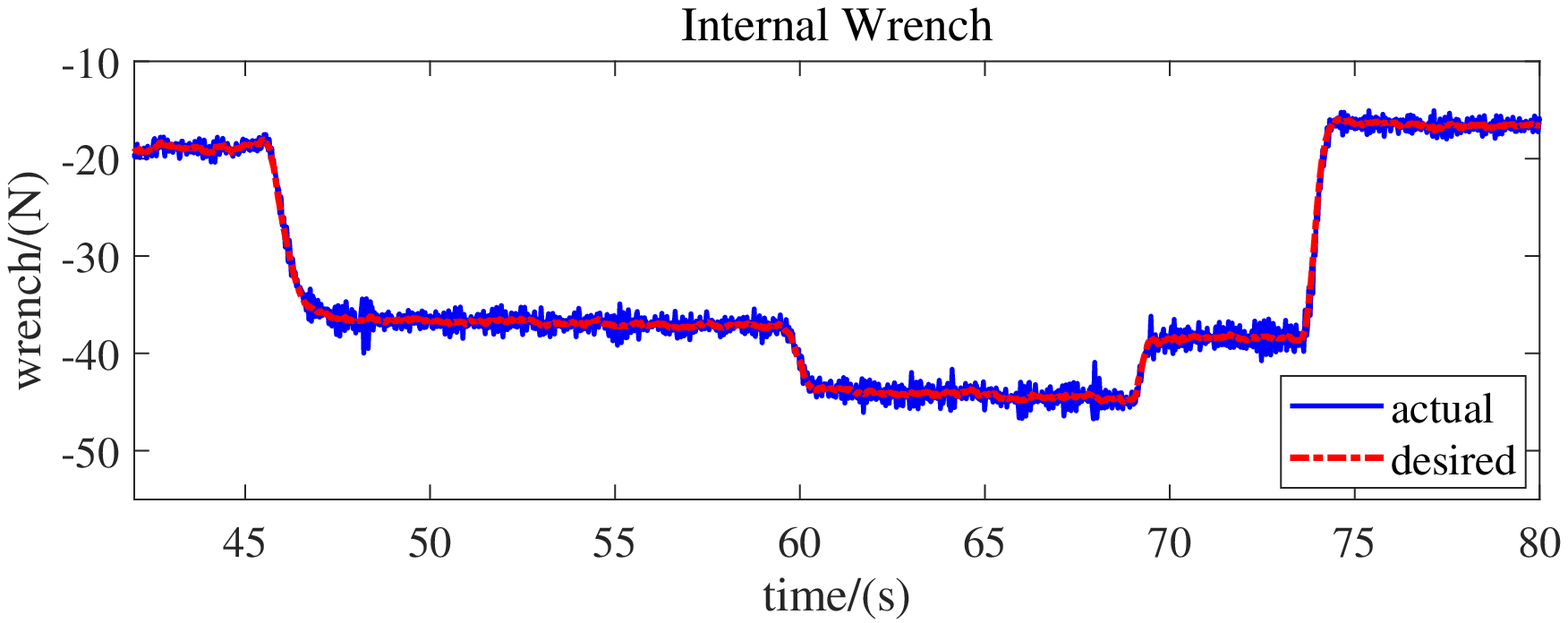}
	}	
	\caption{Actual and desired internal wrench when keeping a constant wrench (a) and tracking an optimized wrench (b). }
	\label{exp2}
\end{figure}

The first experiment demonstrates the advantage of the optimized internal wrench compared with the constant one. The robot clamps a 400g box initially, and then we stack an 800g load and a 200g load on the box in turn. Finally, the desired  \(\boldsymbol{W}^1_{r,dx}\) in (\ref{pid}) is separately set to a constant value 12N shown in Fig.\ref{exp1} (a), and the optimized value obtained by (\ref{optf}) shown in Fig.\ref{exp1} (b).

As shown in Fig.\ref{exp2} (a), the internal wrench is maintained at about 12N during the stable clamping. After stacking an 800g load, the internal wrench is insufficient to provide sufficient friction to resist external disturbance. As a result, the box slips, and the internal wrench returns to 0N. However, in Fig.\ref{exp2} (b), the internal wrench is optimized to a new stable value when stacking an 800g load on it. As we continue to stack up a 200g load, the internal wrench adjusts to a new value again to keep the clamping stable. At this time, the joint torques tend to be saturated, so we stop adding extra loads. The results imply that the optimization approach can adjust the internal wrench according to the change of external disturbance and keep the box clamped stably. 

\begin{figure}[tbhp]
	\centering
	\includegraphics[width=8.5cm]{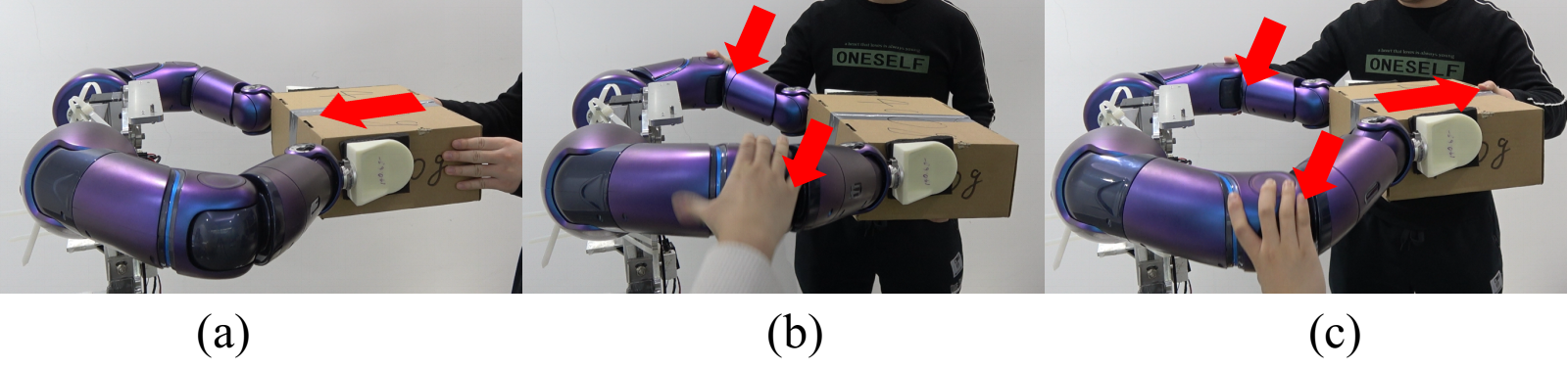}
	\caption{Push and pull the box in any direction (a). Pull the robot arms up and down (b). The two kinds of disturbances are applied simultaneously (c).}
	\label{exp3}
\end{figure}

\begin{figure}[tbhp]
	\centering
	\includegraphics[width=8.5cm]{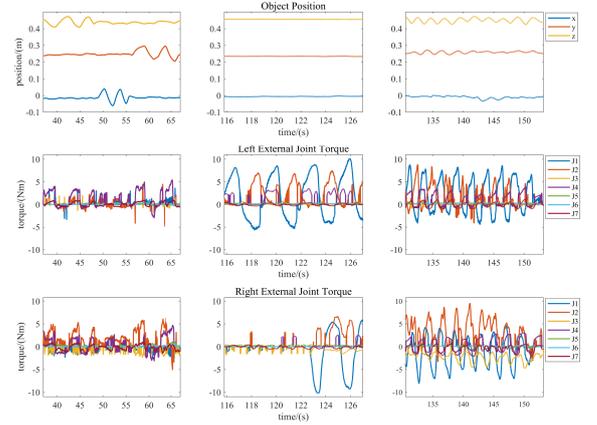}
	\caption{Object position, left and right external joint torques from disturbances on the robot arms. The first column is the time when a human pushes the box. The second column is when disturbances are exerted on the robot arms. And the last one is when disturbances are exerted on both of them at the same time.}
	\label{exp4}
\end{figure}

The second experiment is to verify the ability to resist stochastic external disturbances when keeping stable clamping. The robot clamps a 400g box at the initial moment. Then disturbances are exerted on the box in Fig.\ref{exp3} (a), the robot arm in Fig.\ref{exp3} (b), and both of them simultaneously in Fig.\ref{exp3} (c). During experiments, the orientation of the box  is enforced to keep constant.

The experiment results are shown in Fig.\ref{exp4}. When only interacting with the box along $z$, $x$, and $y$ direction of the base frame, the box responds to the external forces, verifying the object admittance control's effectiveness in the first priority of the proposed control framework. The estimated external joint torques of the left and right arms are small because there is no disturbance on the robot arms. When the disturbance is only exerted on the robot arm, the box's position has nearly no change in any direction. The left and right external joint torques have both large amplitude. It proves that the second priority’s joint space compliance control does not affect the object compliance control in the priority. When the disturbances are exerted on the box and the robot arm simultaneously, they have responses to the external disturbance separately. As a result, it is illustrated that the proposed control approach can realize the compliance control against the object and robot arm’s external disturbances hierarchically while clamping the object stably.
\begin{figure}[!tbhp]
	\centering
	\subfigtopskip=2pt 
	\subfigbottomskip=2pt 
	\subfigcapskip=-2pt 
	\subfigure[]{
		\centering
		\includegraphics[width=3.8cm]{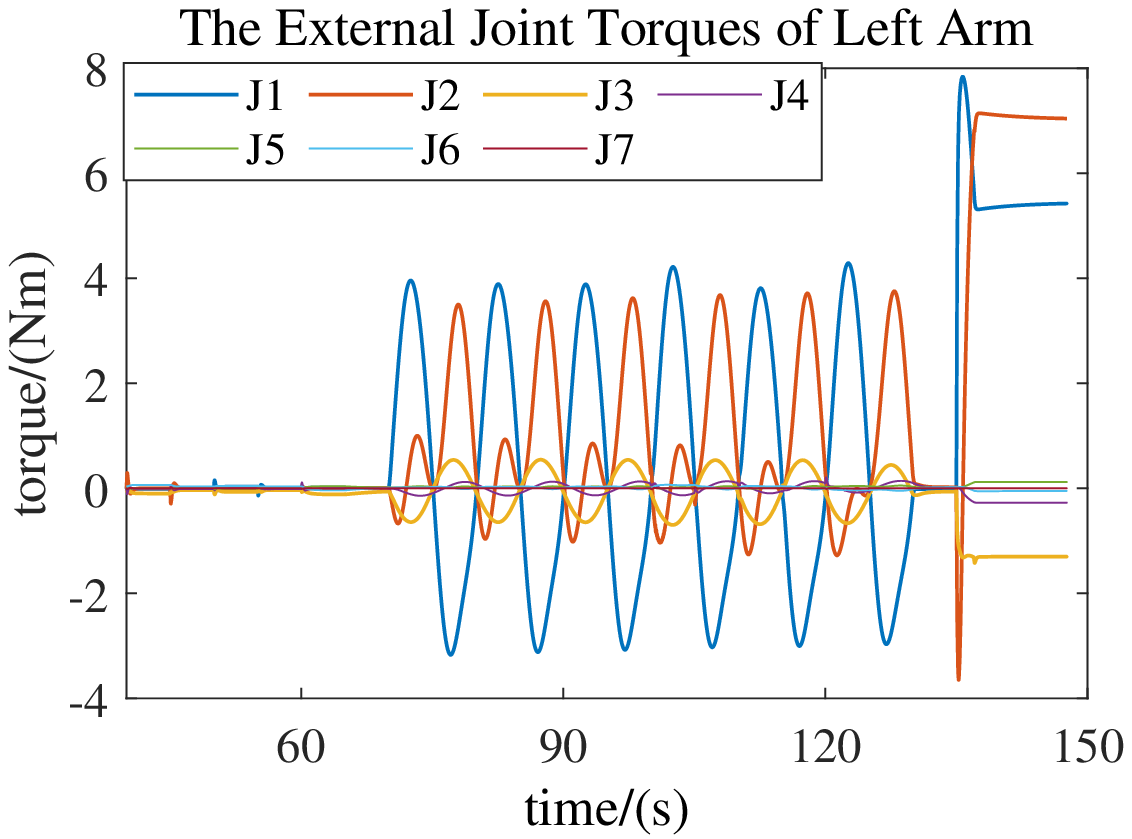}
	}
	\subfigure[]{
		\centering
		\includegraphics[width=3.8cm]{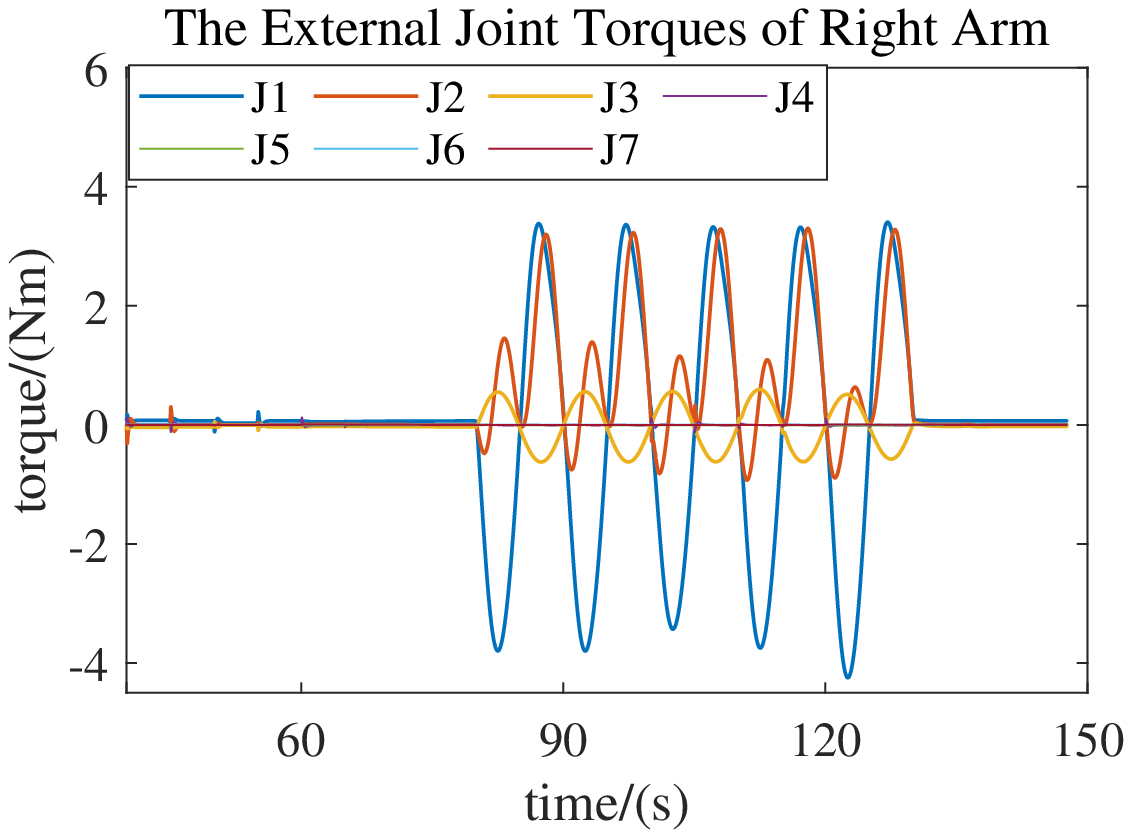}
	}
	\caption{Simulation results: the external joint torques of left arm (a), the external joint torque of right arm (b). }
	\label{simulation_1}
\end{figure}
\begin{figure}[!thbp]
	\centering
	\includegraphics[width=5.0cm]{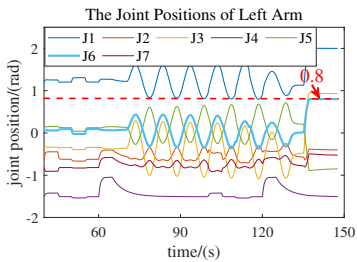}
	\caption{Simulation results: the joint positions of left arm. The red dashed line indicates the upper limit of the sixth joint position.}
	\label{simulation_2}
\end{figure}
	
The second experiment is also performed in the open-source robot simulator Webots\cite{Webots} with ideal frictionless joints. Moreover, a significant external disturbance is exerted on the left arm, in the end, to enforce joints to their limits. Compared with the real robot, the external joint torques shown in Fig.\ref{simulation_1} nearly have no noise. It verifies that the proposed control approach does not cause the noise of the estimated torques shown in Fig.\ref{exp4}. It may be due to errors of friction model and the inaccurate current-to-torque factors. When the external disturbance is applied to the left arm from 125s to 150s, the sixth joint position reaches but dose not exceed its upper limit (0.8 rad) as shown in Fig.\ref{simulation_2}. It illustrates that the proposed control approach can effectively deal with the joint constraints. 

\section{CONCLUSIONS}
We propose a prioritized hierarchical compliance control approach to solve dual-arm clamping control when external disturbances are exerted on the object and the robot arm simultaneously. Firstly, the object and robot arm’s external disturbances are decoupled in the joint space using the momentum observer and the F/T sensors. Then the object admittance and joint space admittance are adopted respectively to realize the compliance control. We optimize the internal wrench to ensure stable clamping under stochastic disturbances according to the friction cone constraints. Finally, the inverse kinematics of the dual-arm robot is solved by hierarchical quadratic programming so that the joint admittance control will not affect the execution of the object control task. The experiments and simulation on the 14-DOF dual-arm robot platform validate the above algorithm’s effectiveness under the two types of external disturbances. However, when the robot clamps an unknown object, it is difficult to preset safety friction coefficients in advance. Therefore, in the future, we will estimate the actual friction coefficients online. 

%
%You may put all reference items in a separate file, say myRef.bib, in bibTex format.
%
%\bibliographystyle{IEEETran}
%\bibliography{myRef} %change to your file name (with suffix .bib)
\bibliographystyle{IEEEtran}
\bibliography{./ROBIO_PHCC}

\begin{thebibliography}{10}
\providecommand{\url}[1]{#1}
\csname url@rmstyle\endcsname
\providecommand{\newblock}{\relax}
\providecommand{\bibinfo}[2]{#2}
\providecommand\BIBentrySTDinterwordspacing{\spaceskip=0pt\relax}
\providecommand\BIBentryALTinterwordstretchfactor{4}
\providecommand\BIBentryALTinterwordspacing{\spaceskip=\fontdimen2\font plus
\BIBentryALTinterwordstretchfactor\fontdimen3\font minus
  \fontdimen4\font\relax}
\providecommand\BIBforeignlanguage[2]{{%
\expandafter\ifx\csname l@#1\endcsname\relax
\typeout{** WARNING: IEEEtran.bst: No hyphenation pattern has been}%
\typeout{** loaded for the language `#1'. Using the pattern for}%
\typeout{** the default language instead.}%
\else
\language=\csname l@#1\endcsname
\fi
#2}}

\bibitem{2018Dual}
K.~Benali, J.~F. Brethe, F.~Guerin, and M.~Gorka, ``Dual arm robot manipulator
  for grasping boxes of different dimensions in a logistics warehouse,'' in
  \emph{2018 IEEE International Conference on Industrial Technology (ICIT)},
  2018.

\bibitem{2017Indirect}
M.~Li, K.~Li, P.~Wang, Y.~Liu, and W.~Guo, ``Indirect adaptive impedance
  control for dual-arm cooperative manipulation,'' in \emph{2017 2nd
  International Conference on Advanced Robotics and Mechatronics (ICARM)},
  2017.

\bibitem{bjerkeng2014fast}
M.~Bjerkeng, J.~Schrimpf, T.~Myhre, and K.~Y. Pettersen, ``Fast dual-arm
  manipulation using variable admittance control: Implementation and
  experimental results,'' in \emph{2014 IEEE/RSJ International Conference on
  Intelligent Robots and Systems}.\hskip 1em plus 0.5em minus 0.4em\relax IEEE,
  2014, pp. 4728--4734.

\bibitem{Ren2016Biomimetic}
Y.~Ren, Y.~Liu, M.~Jin, and H.~Liu, ``Biomimetic object impedance control for
  dual-arm cooperative 7-dof manipulators,'' \emph{Robotics and Autonomous
  Systems}, pp. 273--287, 2016.

\bibitem{murray1994mathematical}
R.~M. Murray, Z.~Li, S.~S. Sastry, and S.~S. Sastry, \emph{A mathematical
  introduction to robotic manipulation}.\hskip 1em plus 0.5em minus 0.4em\relax
  CRC press, 1994.

\bibitem{caccavale2008six}
F.~Caccavale, P.~Chiacchio, A.~Marino, and L.~Villani, ``Six-dof impedance
  control of dual-arm cooperative manipulators,'' \emph{IEEE/ASME Transactions
  On Mechatronics}, vol.~13, no.~5, pp. 576--586, 2008.

\bibitem{heck2013internal}
D.~Heck, D.~Kosti{\'c}, A.~Denasi, and H.~Nijmeijer, ``Internal and external
  force-based impedance control for cooperative manipulation,'' in \emph{2013
  European Control Conference (ECC)}.\hskip 1em plus 0.5em minus 0.4em\relax
  IEEE, 2013, pp. 2299--2304.

\bibitem{2016Coordinated}
L.~Yan, Z.~Mu, W.~Xu, and B.~Yang, ``Coordinated compliance control of dual-arm
  robot for payload manipulation: Master-slave and shared force control,'' in
  \emph{IEEE/RSJ International Conference on Intelligent Robots and Systems},
  2016.

\bibitem{han2019collision}
L.~Han, W.~Xu, B.~Li, and P.~Kang, ``Collision detection and coordinated
  compliance control for a dual-arm robot without force/torque sensing based on
  momentum observer,'' \emph{IEEE/ASME Transactions on Mechatronics}, vol.~24,
  no.~5, pp. 2261--2272, 2019.

\bibitem{tarbouriech2019admittance}
S.~Tarbouriech, B.~Navarro, P.~Fraisse, A.~Crosnier, A.~Cherubini, and
  D.~Sall{\'e}, ``Admittance control for collaborative dual-arm manipulation,''
  in \emph{2019 19th International Conference on Advanced Robotics
  (ICAR)}.\hskip 1em plus 0.5em minus 0.4em\relax IEEE, 2019, pp. 198--204.

\bibitem{chiacchio1996direct}
P.~Chiacchio, S.~Chiaverini, and B.~Siciliano, ``Direct and inverse kinematics
  for coordinated motion tasks of a two-manipulator system,'' 1996.

\bibitem{lin2018projected}
H.-C. Lin, J.~Smith, K.~K. Babarahmati, N.~Dehio, and M.~Mistry, ``A projected
  inverse dynamics approach for multi-arm cartesian impedance control,'' in
  \emph{2018 IEEE International Conference on Robotics and Automation
  (ICRA)}.\hskip 1em plus 0.5em minus 0.4em\relax IEEE, 2018, pp. 1--5.

\bibitem{gao2018projected}
J.~Gao, Y.~Zhou, and T.~Asfour, ``Projected force-admittance control for
  compliant bimanual tasks,'' in \emph{2018 IEEE-RAS 18th International
  Conference on Humanoid Robots (Humanoids)}.\hskip 1em plus 0.5em minus
  0.4em\relax IEEE, 2018, pp. 1--9.

\bibitem{2014Dual}
C.~K. Chou, W.~T. Yang, and P.~C. Lin, ``Dual-arm object manipulation by a
  hybrid controller with kalman-filter-based inputs fusion,'' in
  \emph{Automatic Control Conference}, 2014.

\bibitem{2007Fast}
S.~P. Boyd and B.~Wegbreit, ``Fast computation of optimal contact forces,''
  \emph{IEEE Transactions on Robotics}, vol.~23, no.~6, pp. 1117--1132, 2007.

\bibitem{capurso2017sensorless}
M.~Capurso, M.~M.~G. Ardakani, R.~Johansson, A.~Robertsson, and P.~Rocco,
  ``Sensorless kinesthetic teaching of robotic manipulators assisted by
  observer-based force control,'' in \emph{2017 IEEE International Conference
  on Robotics and Automation (ICRA)}.\hskip 1em plus 0.5em minus 0.4em\relax
  IEEE, 2017, pp. 945--950.

\bibitem{2012Null}
H.~Sadeghian, M.~Keshmiri, L.~Villani, and B.~Siciliano, ``Null-space impedance
  control with disturbance observer,'' in \emph{Intelligent Robots and Systems
  (IROS), 2012 IEEE/RSJ International Conference on}, 2012.

\bibitem{2014Task}
H.~Sadeghian, L.~Villani, M.~Keshmiri, and B.~Siciliano, ``Task-space control
  of robot manipulators with null-space compliance,'' \emph{IEEE Transactions
  on Robotics}, vol.~30, no.~2, pp. 493--506, 2014.

\bibitem{2018Control}
F.~Vigoriti, F.~Ruggiero, V.~Lippiello, and L.~Villani, ``Control of redundant
  robot arms with null-space compliance and singularity-free orientation
  representation,'' \emph{Robotics and Autonomous Systems}, vol. 100, pp.
  186--193, 2018.

\bibitem{flacco2012prioritized}
F.~Flacco, A.~De~Luca, and O.~Khatib, ``Prioritized multi-task motion control
  of redundant robots under hard joint constraints,'' in \emph{2012 IEEE/RSJ
  International Conference on Intelligent Robots and Systems}.\hskip 1em plus
  0.5em minus 0.4em\relax IEEE, 2012, pp. 3970--3977.

\bibitem{flacco2012motion}
F.~Flacco, A.~De~Luca, and O.~Khatib, ``Motion control of redundant robots
  under joint constraints: Saturation in the null space,'' in \emph{2012 IEEE
  International Conference on Robotics and Automation}.\hskip 1em plus 0.5em
  minus 0.4em\relax IEEE, 2012, pp. 285--292.

\bibitem{4058607}
A.~De~Luca, A.~Albu-Schaffer, S.~Haddadin, and G.~Hirzinger, ``Collision
  detection and safe reaction with the dlr-iii lightweight manipulator arm,''
  in \emph{2006 IEEE/RSJ International Conference on Intelligent Robots and
  Systems}, 2006, pp. 1623--1630.

\bibitem{2018A}
C.~Gaz, E.~Magrini, and A.~De~Luca, ``A model-based residual approach for
  human-robot collaboration during manual polishing operations,''
  \emph{Mechatronics}, p. S0957415818300369, 2018.

\bibitem{2008Rigid}
R.~Featherstone, \emph{Rigid Body Dynamics Algorithms}.\hskip 1em plus 0.5em
  minus 0.4em\relax Springer US, 2008.

\bibitem{uchiyama1988symmetric}
M.~Uchiyama and P.~Dauchez, ``A symmetric hybrid position/force control scheme
  for the coordination of two robots,'' in \emph{Proceedings. 1988 IEEE
  international conference on robotics and automation}.\hskip 1em plus 0.5em
  minus 0.4em\relax IEEE, 1988, pp. 350--356.

\bibitem{qpOASES2017}
H.~Ferreau, A.~Potschka, and C.~Kirches, ``{qpOASES} webpage,''
  http://www.qpOASES.org/, 2007--2017.

\bibitem{Webots}
http://www.cyberbotics.com/.

\end{thebibliography}

\end{document}